# Hierarchical Motion Encoder-Decoder Network for Trajectory Forecasting

Qifan Xue, Shengyi Li, Xuanpeng Li, Jingwen Zhao, and Weigong Zhang

Southeast University, Nanjing

`{xue_qifan, li_shengyi, li_xuanpeng, zhao_jingwen, zhangwg}@seu.edu.cn`

## Abstract

*Trajectory forecasting plays a pivotal role in the field of intelligent vehicles or social robots. Recent works focus on modeling spatial social impacts or temporal motion attentions, but neglect inherent properties of motions, i.e. moving trends and driving intentions. This paper proposes a context-free Hierarchical Motion Encoder-Decoder Network (HMNet) for vehicle trajectory prediction. HMNet first infers the hierarchical difference on motions to encode physically compliant patterns with high expressivity of moving trends and driving intentions. Then, a goal (endpoint)-embedded decoder hierarchically constructs multimodal predictions depending on the location-velocity-acceleration-related patterns. Besides, we present a modified social pooling module which considers certain motion properties to represent social interactions. HMNet enables to make the accurate, unimodal/multimodal and physically-socially-compliant prediction. Experiments on three public trajectory prediction datasets, i.e. NGSIM, HighD and Interaction show that our model achieves the state-of-the-art performance both quantitatively and qualitatively. We will release our code here: https://github.com/xuedashuai/HMNet.*

## 1. Introduction

Trajectory prediction is an important function in autonomous systems. Agents' motions depend on their histories, goals, social interactions and constraints from the scene context. Although recent works [1-7] benefit a lot from applying the scene context (e.g. headings, speed, intentions and maps), the context-free method is still worth researching for its generalization and wide range of applications.

A large body of works, including the state-of-the-art methods [4, 8-11], build on Seq2Seq [12] (the encoder-decoder framework). However, they focus on digging the spatial-temporal diversity, but neglect the inherent consistency of the sequence. Widely-used sequence modeling methods such as Long Short-Term Memory (LSTM) [13] and Transformer [14] are first applied on Natural Language Processing (NLP). NLP emphasizes the importance of distinguishing and selecting features among consecutive words (implemented by forget gates and input gates in LSTM) and the long-term dependency (implemented by multi-head attention in Transformer). Unlike NLP, a series of consecutive locations in a trajectory show smooth tendency while words in a sentence are seemingly random. The rough utilization of the Multilayer Perceptron (MLP) layer and the temporal attention mechanism breaks the 'consistency' to some extent. Besides, solely basing on the memory cell (in LSTM) is still not enough to directly learn the sophisticated motion patterns, especially the higher dimensional information like velocity and acceleration. Hence, we propose a novel encoder-decoder architecture which makes a physically compliant decomposition of the motion into location-velocity-acceleration-related patterns.

Undeniably, modeling social interactions is essential for trajectory prediction. However, these social patterns [15, 16] are spatial (mostly depend on relative locations) or contextual [2], but not motional. For example, a distant high-speed (or suddenly braking and turning) agent should have more social impacts than the adjoining but uniformly moving one. Hence, we modify the 'Social Pooling' [16] module for capturing motional social patterns.

Given a partial history, there is no single correct future prediction. Multiple trajectories are plausible and socially-acceptable [2]. Nevertheless, unimodal prediction remains relevant in some situations like the risk assessment. Thus, we employ the CVAE-based [17] goal module to work both unimodally and multimodally.

Our contributions are summarized as follows:

- In order to account for moving trends and driving intensions, we propose a novel context-free encoder-decoder architecture that hierarchically extracts and integrates location-velocity-acceleration-related patterns from trajectories.
- We modify the 'Social Pooling' module as Motional Social Pooling that aggregates integrated motional encodings to model social interactions both physically and socially.
- We demonstrate the state-of-the-art performance and generalization of our HMNet on three public datasets (NGSIM [18, 19], HighD [20] and Interaction [21]), both in unimodal and multimodal cases.

## 2. Related Works

There is a large volume of published studies [22, 23] investigating trajectory prediction in different settings. Conventional methods of trajectory prediction, such as polynomial fitting [24], Gaussian mixture models [25], Kalman Filter [26] and Hidden Markov Models [27], are constrained by hand-crafted rules and features. With the success of the deep neural network, Recurrent Neural Network (RNN)-based approaches like LSTM and Gate Recurrent Unit (GRU) [28] have become prevalent. These works propose to aggregate temporal features in their recurrent hidden states. Recently, the application of transformers [14, 29] achieves a good performance, but its large computational costs should not be ignored.

### 2.1. Contexts and Scene Semantics

Many previous studies import rich contexts, such as the RGB image [1-5] and scene semantics [6, 7, 30]. In the field of vehicle trajectory prediction, road-lanes information [31] and driving states [32] are employed for further improvement. However, the dependency on contexts may decreases the generalization of models because not every scenario can provide adequate contextual clues.

### 2.2. Moving Trends and Driving Intentions

Notably, the trajectory itself is only the observed result of the motion. Accordingly, properties of the motion, such as velocity and acceleration, are worthy of attention. However, only a few studies address this issue. In [33], an asymmetric optimal velocity model is presented to capture the asymmetry between acceleration and deceleration. Christoph Schöller, *et al.* [34] reveal that the constant velocity model performs better than sophisticated models like Sophie [3] and Social GAN [2] in specific scenarios. Xue, *et al.* [35] propose a location-velocity-temporal attention LSTM model but the velocity is only exploited to tweak the final prediction. Furthermore, some researchers [36, 37] introduce driving intentions (overtaking and braking) which could have been proximally portrayed by acceleration. Although direct gaining of velocity and acceleration is fraught with difficulties, we use the hierarchical difference of locations as approximate alternatives.

### 2.3. Social interactions

Behaviors and interactions of surrounding dynamic agents are also crucial cues for the trajectory prediction. Classic approaches propose 'Social Forces' [38, 39] to model interactions as attractive and repulsive forces. Recent methods leverage attention mechanisms [40] or relational reasoning on constructed graphs [41-43] to model interactions. Besides, the 'Social Pooling' [16] method models agents' effects through a novel pooling module. In our work, we apply the 'Social Pooling' module with a few modifications for that its Convolutional Neural Network (CNN) and maxpooling modules have the superior feature aggregation capability.

### 2.4. Unimodality and Multimodality

Early trajectory prediction works focus on unimodal prediction in the form of the concrete trajectory [44] or the probability distribution [15, 16]. Recently, there is a significant focus on multimodal trajectory prediction to capture possible future trajectories given the past. [45, 46] directly predict multiple possible maneuvers and generate corresponding future trajectories given each maneuver. Besides, more works [2, 4] employ generative models such as CVAE [17] and GAN [47] to achieve multimodality. Recent state-of-the-art works [8-10, 48] use goal-conditioned approaches which are regarded as inverse planning or prediction by planning. They learn the final intent or goal of the agent before predicting the full trajectory. Nevertheless, our work has two output modes of both unimodality and multimodality to accommodate different evaluation metrics.

## 3. Methodology

### 3.1. Problem Formulation

In this work, we tackle vehicle trajectory prediction by formulating a probabilistic framework with $N$ interacting agents in the traffic scene. Given past trajectories of $N$ agents as $X = \{X^1, X^2, \cdots, X^n\}$, where $X^n = \{x_{t-\tau}^n, x_{t-\tau+1}^n, \cdots, x_t^n\}$ denotes the $n$-th agent's locations from time $t-\tau$ to $t$ with the past history steps $\tau$. The problem requires predicting future locations $Y = \{Y^1, Y^2, \cdots, Y^n\}$, where $Y^n = \{y_t^n, y_{t+1}^n, \cdots, y_{t+T}^n\}$ denotes locations of all agents from time $t$ to $t+T$ for the future timesteps $T$.

The goal of our work is to learn the probabilistic distribution $p(Y|X)$. On the basis of the classic Seq2Seq [12] model denoted as $p(Y|X) = p(Y|H_s)p(H_s|X)$, we introduce two extra stochastic variables $H_v$ and $H_a$ besides hidden states $H_s$ for locations. $H_v$ and $H_a$ are utilized to account for velocity and acceleration. Formally, we break the distribution into two steps in the encoder-decoder framework as:

$$p(Y|X) = \underbrace{p(Y|H_s, V, X)p(V|H_v, A, X)p(A|H_a, X)}_{\text{Decoder}} \underbrace{p(H_s, H_v, H_a|X)}_{\text{Encoder}}, (1)$$

where $H_s, H_v, H_a$ denote hidden variables of location, velocity and acceleration. $V$ and $A$ represent the prediction of velocity and acceleration. The decoder is composed of three sub-decoders for hierarchically decoding $A$, $V$ and $Y$.

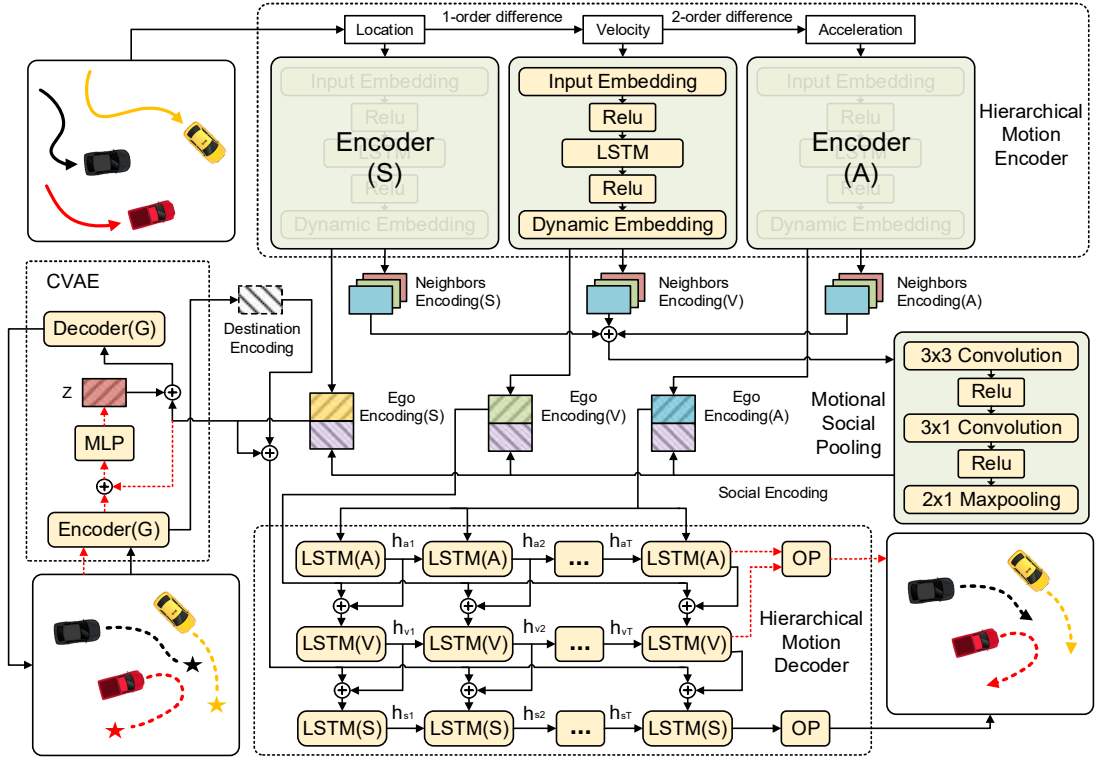

Figure 1. An overview of our HMNet. In the unimodal mode, first we get velocity-related tensors and acceleration-related tensors through 1-order and 2-order difference operations on agents' history trajectories (top-left in the figure). Three types (location-related, velocity-related and acceleration-related) of tensors are encoded separately. Then, we get the social encoding by the Motional Social Pooling module from concatenated neighbors encodings. The three-layers LSTM is set to hierarchically decodes from acceleration to velocity and finally to locations (bottom-right in the figure). In the multimodal mode, the CVAE-based generator (left in the figure) is employed to sample plausible endpoints conditioning the decoder for multimodal prediction. The dotted red lines represent the pipeline only used while training. Further details are in Section 3.2.

Additionally, we employ the CVAE-based goal generator [8] for multimodal trajectory prediction. Goals are defined as the long-term desired destinations which are called as the potential endpoints of trajectories in this paper. In practice, the CVAE [17] samples the latent variable $Z$ to estimate endpoints. Then, the final prediction is conditioned on the estimated endpoint. Hence, we rewrite Eqn. 1 as:

$$p(Y|X) = \underbrace{p(Y|H_s,V,G,X)p(V|H_v,A,X)p(A|H_a,X)}_{\text{Decoder}}$$

$$\cdot \underbrace{p(H_s,H_v,H_a,Z|X)}_{\text{Encoder}} \underbrace{p(G|Z,X)}_{\text{Goal}}. \qquad (2)$$

### 3.2. Model

As shown in Fugure 1, our HMNet consists of four parts: Hierarchical Motion Encoder, Hierarchical Motion Decoder, Motional Social Pooling and the endpoint-based multimodal module. The implement details and code are available in the supplementary material.

#### 3.2.1. Hierarchical Motion Decoder

In reality, location $s$, velocity $v$ and acceleration $a$ can be interconverted by the 1-order and 2-order calculus as:

$$s(t) = \int_0^t v(t)dt = \iint_o^t a(t)dt \qquad (3)$$

or

$$a(t) = \dot{v}(t) = \ddot{s}(t) \qquad (4)$$

That indicates trajectory features being likely implied in these motion properties. Since the discrete sampling of agents' paths, we use the difference as an approximate substitute for the differential. It denotes as:

$$a_t = \frac{v_{t+1} - v_t}{1/f}, v_t = \frac{s_{t+1} - s_t}{1/f}, \qquad (5)$$

where $a_t$, $v_t$ and $s_t$ are acceleration, velocity and location in the frame $t$. $f$ denotes the sampling frequency of the data.

Then, three types of inputs are encoded by the LSTM [13]-based encoder through a rollout approach. The hidden states of the final frame are taken to be the encoding outputs $H_s = (H_s^{ego}, H_s^{nbrs})$, $H_v = (H_v^{ego}, H_v^{nbrs})$ and $H_a = (H_a^{ego}, H_a^{nbrs})$. The joint distribution of the encoder is

$$p(H|X) = p(H_{t-\tau}|X_{t-\tau}) \prod_{i=t-\tau+1}^{t} p(H_i|H_{i-1}, X_i), \quad (6)$$

where $X_t = (s_t, v_t, a_t)$ and $H_t = (H_{st}, H_{vt}, H_{at})$ separately represent sets of inputs and hidden states of the encoder.

### 3.2.2. Motional Social Pooling

Considering the complexity of modeling social interactions, we modify 'Social Pooling' [16] to account for moving trends and driving intensions. We believe that the social attention should not only depends on agents' relative positions, but also be affected by velocity(trends) and acceleration(intensions). Hence, encodings of neighboring agents $H^{nbrs} = cat(H_s^{nbrs}, H_v^{nbrs}, H_a^{nbrs})$ are concatenated to pass through the convolution layers and the maxpooling layer, where $cat(\cdot)$ denotes a concatenation operator. In short, given the neighbors set $\mathcal{N}_i$ of the agent $i$, we construct the social encoding $H^{social}$ by:

$$H^{social} = \phi(\sum_{j \in \mathcal{N}_i} Indiccate(X^i - X^j) H^{nbrs}, W_{social}) \quad (7)$$

where $\phi(\cdot)$ is the 'Social Pooling' operation with weights $W_{social}$. $Indiccate(\cdot)$ represents the indicator function [16] to check if the agent $j$ is the neighbor of the agent $i$.

### 3.2.3. Hierarchical Motion Decoder

In this section, according to the Bayesian principles [49], we suppose that the location $Y$ is independent of the acceleration $A$ if the velocity $V$ is determined:

$$p(Y|V) = p(Y|V, A). \quad (8)$$

Where $V$ depends on $A$ following $p(V|A)$. Hence, we propose the integrated decoder $Dec = (Dec_s, Dec_v, Dec_a)$ which hierarchically predicts $Y$ along with $V$ and $A$. In the first layer, the sub-decoder $Dec_a$ forecasts $A$ by the concatenated encoding $enc_a = cat(H_a, H^{social})$ as:

$$p(A|H_a, X) = p(A_t|enc_a) \prod_{i=t+1}^{t+T} p(A_i|A_{i-1}, enc_a). \quad (9)$$

Then in the second layer, we use the sub-decoder $Dec_v$ and the encoding $enc_v = cat(H_v, H^{social})$ to decode $V$ similarly but considering $A$ as:

$$p(V|H_v, A, X) =$$
$$p(V_t|enc_v) \prod_{i=t+1}^{t+T} p(V_i|V_{i-1}, A_{i-1}, enc_v). \quad (10)$$

Finally, the unimodal trajectory $Y$ is predicted by the sub-decoder $Dec_s$, $V$ and the encoding $enc_s = cat(H_s, H^{social})$ as:

$$p(Y|H_s, V, X) =$$
$$p(Y_t|enc_s) \prod_{i=t+1}^{t+T} p(Y_i|Y_{i-1}, V_{i-1}, enc_s). \quad (11)$$

In practice, $A_i$ and $V_i$ are implemented by the $i$-th hidden states in the recurrent process of the LSTM-based sub-decoder as shown in Figure 1. Especially, all predicted sequences $Y$, $V$ and $A$ are used for supervised learning while only $Y$ is outputted in inference.

### 3.2.4. Endpoint-based Multimodal Module

In order to enable our model with multimodality, we employ the CVAE [17]-based module which generates possible goals (endpoints) $G$ from the latent variable $Z$ and the location encoding $enc_s$. $Z$ is assumed to obey to a standard Gaussian distribution.

In multimodal modes, the sub-decoder $Dec_s$ makes multiple predictions conditioning on estimated endpoints $G$. Hence, we rewrite Eqn. 11 as:

$$p(Y|H_s, V, G, X) =$$
$$p(Y_t|enc_s) \prod_{i=t+1}^{t+T} p(Y_i|Y_{i-1}, V_{i-1}, enc_s, G). \quad (12)$$

### 3.3. Loss Function

We train our HMNet end to end using the following loss function:

$$\mathcal{L}_{HMNet} = \lambda_1 \mathcal{L}_{uni} + \lambda_2 \mathcal{L}_{multi}, \quad (13)$$

$$\mathcal{L}_{uni} = \mathcal{L}_s(Y, \hat{Y}) + \mathcal{L}_v(V, \hat{V}) + \mathcal{L}_a(A, \hat{A}), (14)$$

$$\mathcal{L}_{multi} = D_{KL}(\mathcal{N}(\mu, \sigma) || \mathcal{N}(0, I)) + \mathcal{L}_G(G, \hat{G}), (15)$$

where $Y$, $V$, $A$ and $G$ denote the ground truth while $\hat{Y}$, $\hat{V}$, $\hat{A}$ and $\hat{G}$ are predictions. Losses $\mathcal{L}_s(\cdot)$, $\mathcal{L}_v(\cdot)$, $\mathcal{L}_a(\cdot)$ and $\mathcal{L}_G(\cdot)$ are implemented by Mean Squared Error (MSE) and Negative Log Likelihood as previous work [16]. $D_{KL}(\cdot)$ is the Kullback-Leibler(KL) divergence [50]. In the unimodal mode, $\mathcal{L}_{uni}$ is used. We impose NLL losses directly on the estimated distribution $Y$, $V$ and $A$ rather than on the drawn coordinate samples since the predictions are explicit probability distributions.

In the multimodal mode, $\mathcal{L}_{HMNet}$ is used. We set $\lambda_1 = 1$ and $\lambda_2 = 0.5$ to balance the two losses. The KL divergence term $D_{KL}(\cdot)$ is used for training the CVAE.

## 4. Experiments

In this section, we conduct experiments on three public datasets to access the performance of our HMNet. We follow the widely-used evaluation protocol [2, 8, 9, 16] with 70% data for training, 20% data for validation and 10% data for testing. We observe 15 timesteps (3 seconds) and predict the future 25 timesteps (5 seconds) simultaneously for all agents.

### 4.1. Datasets

**Next Generation Simulation** [18, 19] (**NGSIM**): NGSIM is a collection of video-transcribed datasets of vehicle trajectories on US-101, Lankershim Blvd. in Los Angeles, I-80 in Emeryville. In total, it contains approximately 45 minutes of vehicle trajectory data at 10 Hz and consisting of diverse interactions among cars, trucks, buses and motorcycles in congested flow.

**HighD** [20]: The HighD dataset is a new dataset of naturalistic vehicle trajectories recorded on German highways. Traffic was recorded at six different locations and includes more than 110,500 vehicles. It contains data of 44,500 driven kilometers and 147 driven hours in total. Using state-of-the-art computer vision algorithms, the positioning error is typically less than ten centimeters.

**Interaction** [21]: It is a large-scale real-world dataset which consists of top-down scenes from intersections, highways and roundabouts. The data is collected from three different continents (North America, Asia and Europe). Interaction dataset is challenging as it includes interactions between vehicles, different environments and potentially multiple plausible predictions.

### 4.2. Baseline Models

We compare our HMNet against several published baselines in both unimodal and multimodal modes. Models including previous state-of-the-art methods are briefly described below:
- **LSTM** [12]: The basic LSTM encoder-decoder model.
- **S-LSTM** [15]: Social LSTM uses the fully connected social pooling and generates a unimodal output distribution.
- **CS-LSTM** [16]: Convolutional social LSTM proposes convolutional social pooling to model social influence for vehicle trajectory prediction.
- **S-GAN** [2]: Social GAN is a multi-modal human trajectory prediction GAN trained with a variety loss to encourage diversity.
- **STGAT** [41]: A spatial-temporal graph attention module is proposed to model social interactions.
- **PECNet** [8]: A VAE based state-of-the-art model with goal conditioning predictions.
- **LB-EBM** [9]: A state-of-the-art latent belief energy-based model for diverse human trajectory forecast.
- **HMNet(S)**: The basic configuration of our HMNet dealing with only locations of the trajectory.
- **HMNet(V)**: Extra components are added in the encoder, the decoder and the social pooling module to capture velocity-related patterns.
- **HMNet(V+A)**: The complete form of our HMNet involves all physically complaint patterns including velocity and acceleration.

### 4.3. Evaluation Metrics

The baseline models cover both vehicle and pedestrian prediction methods, yet their evaluation metrics differ. Thus, we adopt the commonly-used ADE/FDE instead of RMSE but set 5 checkpoints at $t = 1s, 2s, 3s, 4s$ and $5s$ as used in vehicle prediction methods. In the multimodal mode, we chose the closest mode to the ground truth from K (K = 20) random samples as in [2, 8, 9]. We still conduct unimodal experiments on NGSIM using RMSE for comparing with the published performance in previous vehicle trajectory prediction works [15, 16]. Three metrics are defined as:

| Metrics | Time | NGSIM (moderate) | | | | | HighD (easy) | | | | | Interaction (hard) | | | | |
|---|---|---|---|---|---|---|---|---|---|---|---|---|---|---|---|---|
| | | LSTM | S-LSTM | CS-LSTM | HMNet (V) | HMNet (V+A) | LSTM | S-LSTM | CS-LSTM | HMNet (V) | HMNet (V+A) | LSTM | S-LSTM | CS-LSTM | HMNet (V) | HMNet (V+A) |
| ADE(m) | 1 s | 0.22 | 0.21 | 0.20 | 0.19 | **0.16** | 0.29 | 0.23 | 0.21 | **0.20** | **0.20** | 0.05 | 0.04 | **0.03** | **0.03** | **0.03** |
| | 2 s | 0.53 | 0.45 | 0.44 | 0.41 | **0.38** | 0.53 | 0.41 | 0.39 | **0.37** | **0.37** | 0.18 | 0.16 | 0.14 | **0.12** | **0.12** |
| | 3 s | 0.93 | 0.72 | 0.70 | 0.66 | **0.62** | 0.78 | 0.60 | 0.57 | **0.55** | **0.55** | 0.41 | 0.39 | 0.35 | 0.31 | **0.30** |
| | 4 s | 1.40 | 1.02 | 1.00 | 0.95 | **0.91** | 1.04 | 0.81 | 0.77 | 0.75 | **0.74** | 0.74 | 0.72 | 0.66 | 0.60 | **0.58** |
| | 5 s | 1.94 | 1.38 | 1.34 | 1.29 | **1.23** | 1.32 | 1.03 | 0.99 | 0.95 | **0.94** | 1.15 | 1.13 | 1.04 | 0.99 | **0.92** |
| FDE(m) | 1 s | 0.42 | 0.38 | 0.37 | 0.35 | **0.31** | 0.48 | 0.37 | 0.35 | **0.33** | **0.33** | 0.11 | 0.09 | 0.08 | **0.06** | **0.06** |
| | 2 s | 1.16 | 0.90 | 0.88 | 0.84 | **0.79** | 0.98 | 0.76 | 0.72 | 0.70 | **0.69** | 0.49 | 0.46 | 0.42 | 0.37 | **0.36** |
| | 3 s | 2.14 | 1.52 | 1.48 | 1.43 | **1.36** | 1.53 | 1.20 | 1.14 | 1.11 | **1.09** | 1.23 | 1.19 | 1.11 | 1.01 | **0.97** |
| | 4 s | 3.35 | 2.30 | 2.23 | 2.16 | **2.08** | 2.14 | 1.68 | 1.61 | 1.56 | **1.53** | 2.29 | 2.26 | 2.07 | 1.99 | **1.86** |
| | 5 s | 4.76 | 3.27 | 3.18 | 3.06 | **2.95** | 2.81 | 2.22 | 2.14 | 2.06 | **2.02** | 3.58 | 3.55 | 3.26 | 3.23 | **2.94** |

Table 1. ADE and FDE metrics on commonly-used datasets in the unimodal mode. The shaded results are the ablation study. Our model outperforms the previous works on both ADE and FDE metrics. The lower the better.

**Average Displacement Error (ADE)**: Average L2 distance between ground truth and our prediction over all predicted time steps.

$$ADE = \frac{\sum_{n \in N} \sum_{t \in T_{pred}} ||\hat{Y}_t^n - Y_t^n||_2}{N \times T_{pred}} \qquad (16)$$

**Final Displacement Error (FDE)**: The distance between the predicted final destination and the true final destination at end of the prediction period $T_{pred}$.

$$FDE = \frac{\sum_{n \in N} ||\hat{Y}_t^n - Y_t^n||_2}{N}, t = T_{pred} \qquad (17)$$

**Root Mean Squared Error (RMSE)**: The root of the mean squared error (MSE) of the predicted trajectories with respect to the true future trajectories.

$$RMSE = \sqrt{\frac{\sum_{n \in N} (\hat{Y}_t^n - Y_t^n)^2}{N}}, t = T_{pred} \qquad (18)$$

## 5. Results and Discussions

In general, three datasets have their own unique scenes. NGSIM and HighD are both collected from highways. Differently, there are highway ramps in NGSIM scenes which lead more lane-changing interactions happening in NGSIM. Meanwhile, Interaction is interaction-rich for its scenes of intersections and roundabouts. In short, the challenging rank of three datasets is: HighD < NGSIM < Interaction.

### 5.1. Quantitate Results

**Unimodality**: In table 1, our model shows a profound improvement compared with the baselines on all ADE/FDE metrics. Even in the most challenging Interaction dataset, our model achieves a significant margin of 11.5% (ADE) and 9.8% (FDE). It suggests that the more interactions, our HMNet shows the better performance with the improvement margin 5.1%/5.6% (HighD), 8.2%/7.2% (NGSIM) and 11.5%/9.8% s(Interaction). It can thus be concluded that our model benefits from the physically-compliant patterns of agents' motions. In addition, the RMSE results shown in Table 2 proves the effectiveness of our work by the improvement margin of 7.6%.

|  | Time | LSTM* | S-LSTM* | CS-LSTM* | HMNet (V) | HMNet (V+A) |
|---|---|---|---|---|---|---|
| RMSE(m) | 1 s | 0.68 | 0.65 | 0.64 | 0.53 | **0.50** |
|  | 2 s | 1.65 | 1.31 | 1.27 | 1.18 | **1.13** |
|  | 3 s | 2.91 | 2.16 | 2.09 | 1.96 | **1.89** |
|  | 4 s | 4.46 | 3.25 | 3.10 | 2.95 | **2.85** |
|  | 5 s | 6.27 | 4.55 | 4.37 | 4.17 | **4.04** |

Table 2. RMSE metrics on NGSIM in the unimodal mode. The performance of models (denoted by *) is published in [15, 16]. The shaded results are the ablation study. Our model achieves the best RMSE. The lower the better.

**Multimodality**: As is shown in Table 3, models' performance in the multimodal mode deserves discussion by datasets.

In NGSIM dataset, we notice that the best FDE (0.43m) is achieved by PECNet while its rest FDEs are not very encouraging. The undulating performance of PECNet is expected as in [8] for focusing on optimizing the goal or the final step. The phenomenon somehow happens in LB-EBM for the same sake. Nevertheless, our model achieves the best ADE with the improvement margin of 6.9% compared with LB-EBM.

Considering high-linear trajectories in HighD, PECNet and LB-EBM have a disappointing performance even compared with S-GAN and STGAT. It can be assumed that the LSTM layers (in S-GAN and STGAT) are suitable for linear highway trajectory prediction compared with the MLP layers (in PECNet and LB-EBM). However, our HMNet still has the best ADE of 0.22m.

| Dataset | Time | S-GAN | STGAT | PECNet | LB-EBM | HMNet(S) | HMNet(V) | HMNet(V+A) |
|---|---|---|---|---|---|---|---|---|
| NGSIM (moderate) | 1s | 0.19 / 0.33 | 0.28 / 0.47 | 0.28 / 0.46 | 0.22 / 0.41 | 0.22 / 0.37 | 0.20 / 0.35 | **0.13 / 0.27** |
|  | 2s | 0.34 / 0.73 | 0.58 / 1.16 | 0.50 / 0.87 | 0.39 / 0.64 | 0.40 / 0.72 | 0.38 / 0.69 | **0.37 / 0.67** |
|  | 3s | 0.57 / 1.23 | 0.95 / 2.06 | 0.68 / 1.16 | 0.54 / 0.88 | 0.55 / 0.92 | 0.52 / 0.89 | **0.51 / 0.87** |
|  | 4s | 0.83 / 1.79 | 1.39 / 3.13 | 0.82 / 1.29 | 0.66 / 0.97 | 0.65 / 0.94 | 0.62 / 0.95 | **0.61 / 0.92** |
|  | 5s | 1.21 / 2.55 | 1.88 / 4.35 | 0.88 / **0.43** | 0.72 / 0.63 | 0.70 / 0.91 | 0.69 / 1.01 | **0.67** / 0.98 |
| HighD (easy) | 1s | 0.07 / 0.12 | 0.07 / 0.11 | 0.20 / 0.37 | 0.22 / 0.41 | 0.07 / 0.11 | 0.06 / 0.09 | **0.05 / 0.07** |
|  | 2s | 0.13 / 0.25 | 0.12 / 0.23 | 0.49 / 1.11 | 0.38 / 0.64 | 0.12 / 0.22 | 0.10 / 0.20 | **0.08 / 0.15** |
|  | 3s | 0.20 / 0.39 | 0.18 / 0.35 | 0.82 / 1.77 | 0.54 / 0.87 | 0.18 / 0.34 | 0.16 / 0.31 | **0.12 / 0.25** |
|  | 4s | 0.27 / 0.55 | 0.24 / 0.48 | 1.06 / 1.77 | 0.66 / 0.96 | 0.24 / 0.49 | 0.21 / 0.44 | **0.17 / 0.38** |
|  | 5s | 0.38 / 0.78 | 0.31 / 0.63 | 1.08 / **0.41** | 0.72 / 0.62 | 0.30 / 0.68 | 0.27 / 0.65 | **0.22** / 0.52 |
| Interaction (hard) | 1s | 0.56 / 0.98 | 0.17 / 0.28 | 0.27 / 0.46 | 0.25 / 0.43 | 0.09 / 0.18 | **0.04 / 0.08** | **0.04** / 0.07 |
|  | 2s | 0.99 / 1.71 | 0.36 / 0.72 | 0.52 / 1.07 | 0.51 / 1.01 | 0.26 / 0.65 | 0.14 / 0.39 | **0.13 / 0.36** |
|  | 3s | 1.39 / 2.51 | 0.62 / 1.39 | 0.87 / 2.04 | 0.78 / 1.56 | 0.52 / 1.39 | 0.33 / 1.02 | **0.31 / 0.97** |
|  | 4s | 1.80 / 3.35 | 0.95 / 2.25 | 1.32 / 3.33 | 1.12 / 2.68 | 0.85 / 2.37 | 0.61 / 1.92 | **0.58 / 1.85** |
|  | 5s | 2.40 / 4.40 | 1.34 / 3.29 | 1.77 / 3.37 | 1.48 / 3.28 | 1.25 / 3.52 | 0.97 / 3.04 | **0.93 / 2.94** |

Table 3. ADE/FDE metrics (in meters) on commonly-used datasets in the multimodal mode. We randomly sample K (K = 20) predicted trajectories and choose the minimum ADE/FDE as the evaluation metrics. The shaded results are the ablation study. Our model outperforms state-of-the-art methods and especially works in interaction-rich scenes.

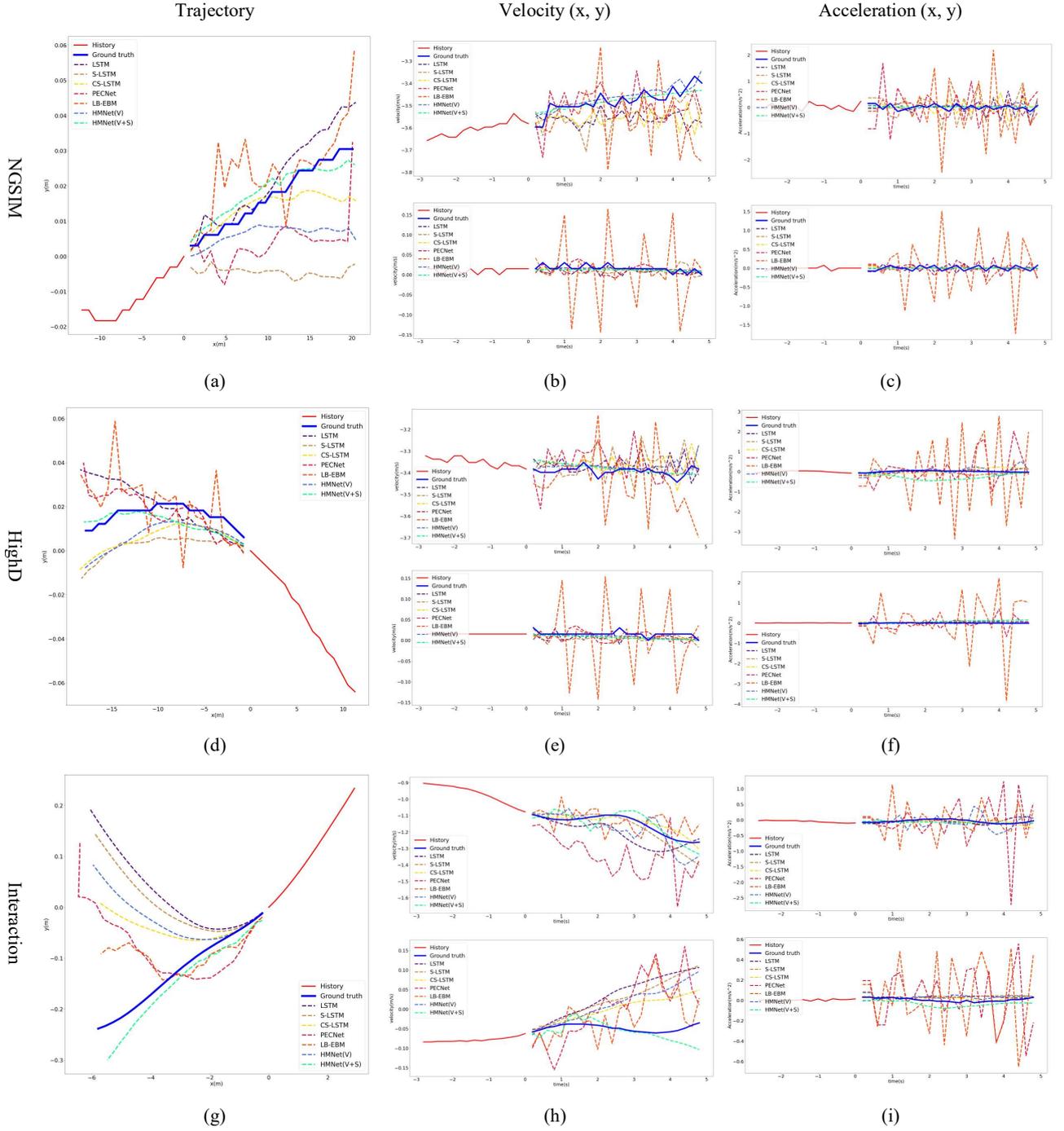

Figure 2. Qualitative results of models across three datasets. The first column presents the predicted trajectory where some paths are sawtooth-like due to the scaling of axis. The second column is results of the velocity. We present the velocity in x-axis and y-axis separately. The third column presents the acceleration where we use the same illustration method as the velocity.

In the most challenging Interaction dataset, our model achieves the state-of-the-art performance with the improvement margin of 37.1% (ADE) and 10.4% (FDE). This result may support the hypothesis that that our model digs the physically-complaint patterns to take the moving trends and driving intentions into account.

**Ablation study**: We conduct ablation studies as shown in Table 1, 2, 3 (shaded results) with three configurations to examine the components of our model. It is noticed that we didn't test HMNet(S) in the unimodal mode because HMNet(S) has the exactly same components (LSTM and 'Social Pooling') and the matching performance with CS-LSTM. Instead, we examine the component of adding velocity-related patterns into the encoder, the decoder and the social pooling module as HMNet(V). HMNet(V) achieves the best improvement margin of 22.4% (ADE) and 13.6% (FDE) compared with HMNet(S). Additionally, HMNet(V+A) takes a step further than HMNet(V) with the best improvement margin of 7.1% (ADE) and 3.3% (FDE). This indicates that driving intensions (acceleration-related patterns) is of same importance as moving trends and our model can extract and utilize it efficiently.

## 5.2. Qualitative Results

We illustrate qualitative results in Figure 2. In the first column, it is noticed that the MLP-based models (PECNet and LB-EBM) predict sawtooth trajectories while those of LSTM-based models (our HMNet and CS-LSTM) are smooth. This can therefore be assumed that the MLP layer independently predicts locations with different weights while LSTM has the memory cell. These two structures each has its pros and cons. The trajectory of LSTM looks plausible but it sometimes loses its way as in Figure 2(g). In contrast, trajectories of PECNet and LB-EBM are flexible and they can make sharp turnings due to the goals as in Figure 2(a).

In terms of velocity and acceleration, our model outperforms others to make more accurate and plausible predictions as shown in the second and third columns. Especially, the swaying prediction of LB-EBM can thus be suggested that it uses 4 sub-goals in inference as in [9].

## 5.3. Limitations

We set up the neighboring agents by defining a field of the ego agent covering 60m in length and 3 lanes in width, as in [16]. We apply the setting on all baseline models. Agents outside the field are not considered as the neighbors. This may cause that we omit the information of agents separated by the lane and ultra-long-distance agents. The assumption is pointed out because it may become the potential limitation.

## 6. Conclusion

In this work, we present HMNet, a novel context-free Hierarchical Motion Encoder-Decoder Network for vehicle trajectory forecasting. HMNet is able to encode motions in the form of physically compliant patterns using the hierarchical motion difference. Further, the goal (endpoint)-based decoder can construct unimodal/multimodal prediction through the location-velocity-acceleration chain. We also modify the 'Social Pooling' module as Motional Social Pooling which captures moving trends and driving intensions to represent social interactions. We benchmark HMNet across three datasets including NGSIM, HighD and Ineraction. Our model achieves the state-of-the-art performance both quantitatively and qualitatively.

This study is limited due to the lightweight memory cell in LSTM. Further work needs to be done to establish a sophisticated memory architecture which can learn more information from motion patterns.